\documentclass[10pt,twocolumn,letterpaper]{article}

\usepackage[pagenumbers]{cvpr} 

\usepackage[dvipsnames]{xcolor}

\usepackage{amssymb}
\usepackage{pifont}
\usepackage{xcolor,colortbl}
\definecolor{green}{rgb}{0.1,0.1,0.1}

\usepackage{mathtools}

\definecolor{cvprblue}{rgb}{0.21,0.49,0.74}
\usepackage[pagebackref,breaklinks,colorlinks,citecolor=cvprblue]{hyperref}

\def\paperID{5140} 
\def\confName{CVPR}
\def\confYear{2024}

\title{Bootstrapping Autonomous Driving Radars with Self-Supervised Learning}

\author{
	\makebox[.25\linewidth]{Yiduo Hao\thanks{Work done during internship at EPFL.}~\textsuperscript{ \textdagger}}\\ University of Cambridge
\and
\makebox[.25\linewidth]{Sohrab Madani\thanks{denotes co-primary first authors.}}\\ UIUC
\and
\makebox[.25\linewidth]{Junfeng Guan}\\ EPFL \vspace{6pt}
\and
\makebox[.25\linewidth]{Mohammed Alloulah\thanks{Work done whilst at Nokia Bell Labs.}}\\ RadarEye
\and
\makebox[.25\linewidth]{Saurabh Gupta}\\ UIUC
\and
\makebox[.25\linewidth]{Haitham Hassanieh}\\ EPFL
}

\newcommand{\name} {{\it Radical}}
\newcommand{\Radatron} {{\it Radatron}}

\begin{document}
\maketitle
\begin{abstract}
The perception of autonomous vehicles using radars has attracted increased research interest due its ability to operate in fog and bad weather. However, training radar models is hindered by the cost and difficulty of annotating large-scale radar data. To overcome this bottleneck, we propose a self-supervised learning framework to leverage the large amount of unlabeled radar data to pre-train radar-only embeddings for self-driving perception tasks. 
The proposed method combines radar-to-radar and radar-to-vision contrastive losses to learn a general representation from unlabeled radar heatmaps paired with their corresponding camera images. 
When used for downstream object detection, we demonstrate that the proposed self-supervision framework can improve the accuracy of state-of-the-art supervised baselines by $5.8\%$ in mAP. Code is available at \url{https://github.com/yiduohao/Radical}.
\end{abstract}
\section{Introduction}
\label{sec:intro}
Millimeter-wave (mmWave) radars have received increased interest from the self-driving cars industry owing to its cost-effectiveness and its ability to operate in adverse weather conditions where cameras and lidar fail
like in fog, smog, snowstorms, and sandstorms~\cite{zang2019impact,norouzian2019snowfall,norouzian2020rain}.
As such, there has been a significant amount of work, from both academia~\cite{guan2020through,wang2021rethinking,barnes2020oxford,sheeny2020radiate} and industry~\cite{rebut2022radial,mostajabi2020high,ouaknine2021carrada,mostajabi2020zendar}, on developing data-driven methods for semantic scene understanding on top of radar signals. Moreover, the advent of standard commercial automotive radars has made real-world deployments and large-scale data collection campaigns possible and several automotive radar datasets have recently been curated~\cite{caesar2020nuscenes, ouaknine2021carrada, wang2021rethinking, barnes2020oxford, sheeny2020radiate, mostajabi2020high, nowruzi2020deep, zhang2021raddet, madani2022radatron}.

\begin{figure}[t]
    \begin{center}
        \includegraphics[width=\columnwidth]{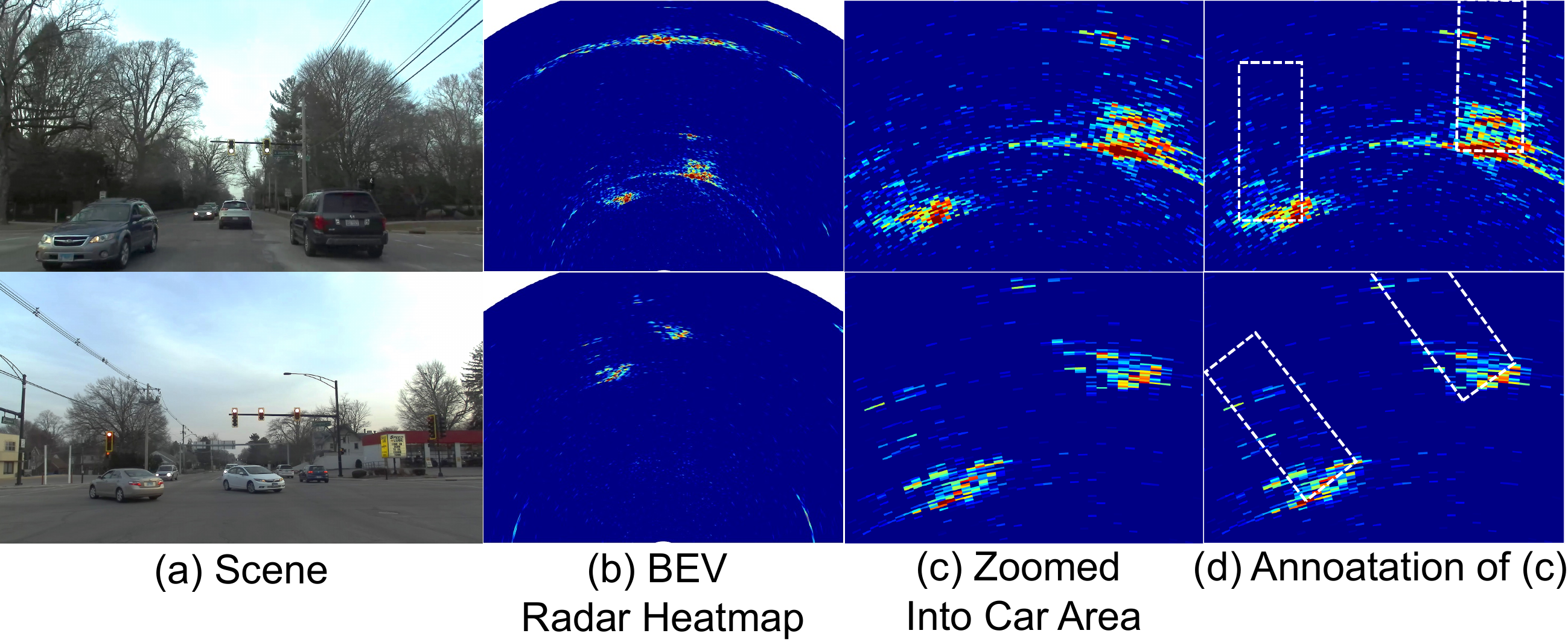}
        \vskip -0.1in
        \caption{\textbf{Millimeter wave radar heatmaps are uninterpretable to humans and are hence difficult to annotate.}}
        \label{fig:label_challenge}
    \end{center}
    \vskip -0.3in
\end{figure}

However, compared to de facto computer vision datasets like ImageNet, the volume of annotated open radar datasets remains very limited. This is because radar images are especially challenging for humans to interpret and thus annotate. Figure~\ref{fig:label_challenge} shows an example of bird's eye view (BEV) radar heatmaps and the corresponding camera images.  Unlike camera images, radar heatmaps appear as blobs with no sharp boundaries or well-defined shapes for the objects present in the scene. These blobs carry little to no contextual or perceptual information and, as such, are hard to interpret by humans. Furthermore, mmWave radar signals are highly specular; meaning, mmWave signals exhibit mirror-like reflections on cars~\cite{bansal2020pointillism}. 
As a result, not all reflections from the car propagate back to the radar receiver, and most of the car does not appear in the image. These effects compound making it difficult even for well-trained radar imaging experts to draw precise bounding boxes of objects~\cite{guan2020through}. As a result, only a tiny fraction of radar data is typically labeled (e.g. 10\%) of the hundreds of thousands of raw radar frames in open radar datasets~\cite{madani2022radatron}. Hence, building accurate supervised radar object detection models is extremely difficult.

To address the challenge of annotating radar data, prior work leverages other sensing modalities like cameras and lidar to derive labels for radar heatmaps and use these labels as groundtruth to train radar-based models~\cite{orr2021high,wang2021RODNet,weston2019probably, sless2019road, kung2022radar,kaul2020rss,ding2023hidden}. However, different sensory modalities have different viewpoints and projection planes of the scene. For example, camera-based labels suffer from depth-unaware perspective projection onto the image plane, so they cannot provide accurate supervision along the depth axis in BEV radar heatmaps. Errors in viewpoint alignment between the different sensory modalities also result in highly inaccurate detection. Moreover, because radar and optical sensors (camera and lidar) operate on orthogonal portions of the electromagnetic spectrum, objects that are visible to optical sensing are not necessarily visible to radar and vice versa. Directly using lidar data to supervise the training of radar will force the radar model to focus too much on less prominent reflections in radar heatmaps, such as less-visible surfaces due to specularity. In contrast, certain materials, such as glass, are not visible to optical sensors but are visible to radars. Therefore, cross-modal supervision results in false positive and false negative detection~\cite{kung2022radar}. 
Finally, as radar hardware continues to evolve, it requires us to keep labeling new datasets collected using new radar hardware, which is going to be very expensive in the long run.


In this paper, we aim to leverage large-scale unlabeled radar data but bypass the complexities of explicit annotations.
We propose a self-supervised learning approach that uses a joint embedding architecture to pre-train a radar object detector using distillation from vision and radar itself.
Learning under our cross-modal and intra-modal objectives happens at the mutual information level~\cite{oord2018representation,alloulah2022self}, rather than explicitly annotating radar data as in prior work~\cite{orr2021high,wang2021RODNet,weston2019probably, sless2019road, kung2022radar,kaul2020rss,ding2023hidden}.

Applying self-supervised learning (SSL), which has been extensively studied in the NLP and CV communities, to the radar domain, is nontrivial because state-of-the-art self-supervised learning methods are designed for camera images. They either design pretext prediction tasks for RGB images~\cite{devlin2018bert,he2022masked}, or leverage camera-specific attributes to design strong augmentations to enforce semantic invariance~\cite{chen2020simple, chen2021exploring, pu2023finegrained}.
RGB augmentation methods cannot be generalized to RF sensing data, including radar. For example, radar data are natively associated with polar coordinates and hence are not invariant to transformations like translation and resizing. 
Previous work~\cite{li2022unsupervised} on sensing the human pose found that directly applying popular SSL frameworks like~\cite{chen2020simple, he2020momentum, wu2018unsupervised} to radar heatmaps results in ``shortcuts'' in the learnt representation rather than capturing meaningful radar information.

We address these challenges by presenting \name, a radar-based object detection system, that is fine-tuned on top of pre-trained radar embeddings to accurately estimate object bounding boxes from radar alone, e.g., during a snowstorm when vision and lidar fail. Our contributions are threefold: 
\begin{itemize}
    \item First, we propose a new contrastive learning framework using radar heatmaps and vision.
It combines both cross-modal (radar-to-vision) and intra-modal (radar-to-radar) contrastive loss terms. The cross-modal term allows us to distill priors from vision such as object semantics in self-driving environments and the intra-modal term allows us to distill priors underlying radar structure such as sparsity and specularity. 

    \item Second,  we introduce a novel augmentation technique RMM (Radar MIMO Mask) that is tailored for state-of-the-art automotive radars. RMM leverages the fact that these radars use MIMO which combines multiple transmitters and multiple receivers. We manipulate how we combine the raw signals coming from different transmitter/receiver pairs to generate new augmented radar heatmaps. This augmentation preserves the underlying geometric structure of the scene while mimicing the radar noise induced by Doppler phase distortions~\cite{guan2023exploit}.

    \item Third, we conduct extensive evaluations and demonstrate significant improvements in radar-only 2D bounding box detection using our framework. Specifically, our results show that \name\ improves the mean average precision (mAP) metric of car detection by 5.8\% compared to supervised learning.

\end{itemize}

 To the best of our knowledge, this is the first work on autonomous driving that uses self-supervised learning to take advantage of the vast amounts of unlabeled radar data and achieve 2D bounding box detection using radar only. Our findings may prove key in generating pre-trained models that avoid the need to annotate massive amounts of radar data and enable lifelong learning on new radar hardware and datasets.

\begin{figure*}[t]
	\begin{center}
		\includegraphics[width=\textwidth]{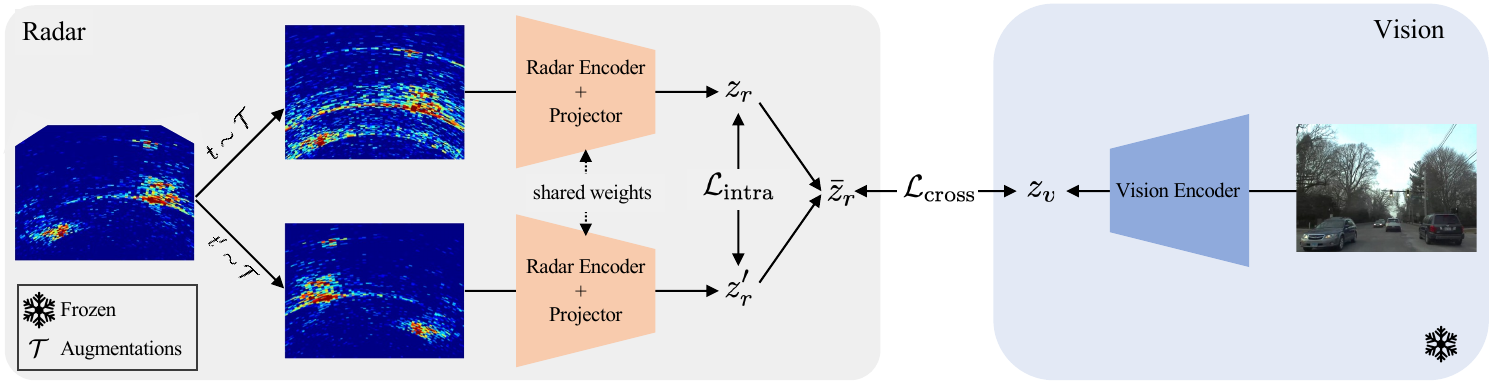}
		\vskip -0.1in
		\caption{\textbf{Overall network of \name.} Knowledge is distilled from a pretrained vision model into a radar model. A mini-batch of $B$ radar-vision pairs flow through network, whose encodings interact locally within the radar branch and globally across the radar and vision branches. That is, \name~is trained using a composite contrastive loss with \emph{intra-} and \emph{cross-modal} terms. }
		\label{fig:overall}
	\end{center}
	\vskip -0.3in
\end{figure*}

\section{Related Work} \label{sec:related_work}

\noindent \textbf{Self-supervised learning.} 
SSL, in its contrastive and non-contrastive flavours, has by now become a staple of representation learning for computer vision tasks~\cite{chen2020simple,he2020momentum,grill2020bootstrap,zbontar2021barlow,oord2018representation,caron2020unsupervised,bardes2021vicreg,goyal2022vision, ma2023revisiting}.
At the core of vision SSL lies augmentation for synthetically generating positive views for enforcing semantic invariance.
 We build on two pioneering contrastive SSL methods for vision: SimCLR and MoCo~\cite{chen2020simple,he2020momentum,chen2020improved}.
SimCLR introduced the canonical contrastive architecture using in-batch negative sampling, which typically relies on a large batch size and associated memory.
MoCo uses an efficient queue and momentum update, which decouples negative sampling from the batch size. Although we heavily draw on vision SSL, our work recasts recent advances within a new cross-modal learning objective for accurate vision-free bounding box estimation.

\vskip 0.05in
\noindent \textbf{Cross-modal SSL.} 
SSL's earlier NLP breakthroughs along with recent vision successes have spawned a plethora of new methods tackling representation learning under multi-modal settings~\cite{baltruvsaitis2018multimodal}, whereby 
paired positive views from other modalities replace or complement augmentation in vision SSL.
Examples include vision and sound~\cite{alwassel2020self,arandjelovic2017look,arandjelovic2018objects,asano2019self,aytar2016soundnet,morgado2020learning,owens2016ambient,afouras2020self}, vision and text~\cite{radford2021learning,jia2021scaling}, different formats of medical imaging~\cite{windsor2021self}, vision and point clouds~\cite{afham2022CrossPoint,huang2021spatio,xie2020pointcontrast}, and vision and radar~\cite{jain2022multimodal,prexl2023multi,alloulah2023look,alloulah2022self}.
Our work expands on the early literature of radio-visual SSL, and further addresses the peculiarities of practical automotive radar, i.e., differs drastically from satellite-mounted radar for remote sensing~\cite{jain2022multimodal,prexl2023multi} while achieving accurate radio-only bounding box car detection, as opposed to simple scene classification in~\cite{alloulah2022self} or label-free target localization (center only) in~\cite{alloulah2023look}.

\vskip 0.05in
\noindent \textbf{Radio SSL.}
An emerging body of literature treats SSL for radio signals such as radar and WiFi~\cite{li2022unsupervised, song2022rf, xiang2023MAE, cao2021towards, yang2022unsupervised}.
Radio signals represent another SSL data domain~\cite{balestriero2023cookbook} that comes with a unique set of challenges and considerations.
Despite some early prior work~\cite{li2022unsupervised,song2022rf}, there remain no mature recipes for data augmentation in the radio domain.
For instance, Li et al. demonstrate that the naive application of popular contrastive learning methods to radio signals gives rise to \emph{shortcuts} in the learned representation, and propose radio-specific transformations in mitigation~\cite{li2022unsupervised}.
Similarly, RF-URL~\cite{song2022rf} employs signal processing techniques, specific to each WiFi and radar data formats, for augmentation in order to use these radio signals within popular SSL architectures.
Our cross-modal work differs from radio-only SSL literature because we also rely on vision which we argue brings robustifying and constraining priors to the much sparser radio domain.
Our composite SSL loss, however, does similarly contain a radio-only term for which we devise a new augmentation scheme that we extensively characterize and benchmark.

\section{Background on mmWave Radar} 
\label{sec:background}

Millimeter-wave radars transmit FMCW (Frequency Modulated Continuous Wave) radar waveforms and receive the reflections off objects in the environment to estimate the round-trip Time-of-Flight (ToF) $\tau$, and hence the ranges of the reflectors $\rho = \tau c/2$ (c denotes the speed of light) in the scene. Furthermore, to localize objects in the 2D range-azimuth polar coordinate $(\rho,\phi)$ and create a 2D bird's eye view radar heatmap, we need to use multiple receiver (RX) antennas to capture the minute ToF differences $\Delta\tau_{ij}=\tau_i-\tau_j$ between different RX. It allows us to estimate the azimuth angle ($\phi$) from which the reflections arrive ~\cite{iovescu2017fundamentals}.

However, to be viable for semantic scene understanding and object detection, we must overcome the resolution limitations of radar along with a number of unique challenges.
Although the wide bandwidth of mmWave radars allows us to achieve a cm-level ranging resolution, the angular resolution is bounded by the number of antenna elements and the antenna aperture size.
Fortunately, the recent innovation of cascaded MIMO radars provides a much more scalable solution. It uses N TX and M RX \textit{physical} antennas to emulate N$\times$M \textit{virtual} antenna links. This allows the angular resolution to scale bilinearly with the number of antennas, even though the resulting angular resolution is still no where near those of cameras and lidars. 

Nevertheless, cascaded MIMO radars suffer from motion smearing in highly dynamic scenes, such as moving cars on the road, due to {\it Doppler-induced phase noise}~\cite{madani2022radatron,guan2023exploit}. Consequently, radar reflections can become smeared and even appear at completely different locations.
Moreover, unlike optical signals, mmWave signals are highly specular, that is, signals exhibit mirror-like reflections on cars~\cite{reina2011radar}. As a result, not all reflections from the car propagate back to the mmWave receiver, and most of the car does not appear in the image, making it impossible to detect its shape~\cite{guan2020through}.

Finally, radar heatmaps appear as blobs with no sharp boundaries or shapes of objects, where the voxel values represent per-voxel reflected signal energy from objects in the scene. Therefore, radar heatmaps carry little to no contextual and perceptual information and are difficult for humans to interpret and annotate.



\section{Method}
\label{sec:technical}




Our primary goal is to pretrain a radar backbone net on large-scale data in a self-supervised fashion.
The learnt radar embeddings can then be employed in various downstream tasks.
To achieve this goal, we build an SSL framework that feeds on both standalone radar and paired radar-vision data.
Specifically, our \name~net implements a composite SSL loss with two terms: (a) intra-modal, and (b) cross-modal. 
The intuition is that the radar-to-radar intra-modal loss term focuses on structures specific to radar data, as we explain further in Secs.~\ref{sec:intra}~\&~\ref{sec:augs}.
The radar-to-vision cross-modal term, on the other hand, learns structures of scenes on the road where visual priors play an important role in constraining and robustifying the features of the sparser radar modality.
By employing both intra-modal and cross-modal SSL, the network feeds on unlabeled radar-vision data to learn a powerful radar representation which works well on a car detection downstream task, as we demonstrate in Sec.~\ref{sec:exp}.
In the remainder of this section, we explain each loss term in more detail.

\subsection{Distillation setup}

Let $(r, v) \in \mathcal{D}$ be a radar-vision data pair in dataset $\mathcal{D}$, 
where $r \in \mathbb{R}^{1 \times L \times A}$ is a radar heatmap with $L$ range bins and $A$ azimuth bins, and $v \in \mathbb{R}^{3 \times H \times W}$ is a corresponding RGB image. 
Encode the radar heatmap with a backbone net $f_{\theta^r}$ then project it with an MLP head $g_{\phi^r}$, assuming some weight parametrisation $\{\theta^r, \phi^r\}$, such that $z_r = g_{\phi^r}(f_{\theta^r}(r)) \in \mathbb{R}^{N}$.
Similarly encode the paired visual image such that $z_v = f^{\ast}_{\theta^v}(v) \in \mathbb{R}^{N}$, with $f^{\ast}_{\theta^v}$ being a pretrained and frozen vision backbone model.
Knowledge is distilled from the pretrained vision backbone $f^{\ast}_{\theta^v}$ and into the radar model $f_{\theta^r}$ by means of local interactions at the radar branch, as well as global interactions with the vision branch as depicted in Fig.~\ref{fig:overall}.  

\subsection{Intra-modal radar learning}
\label{sec:intra}

For radar, we aim to enrich the learnt embeddings with attributes that would enhance their discriminative power and robustness.
To this end, we design a set of augmentations $\mathcal{T}$ (cf., Sec.~\ref{sec:augs}) and formulate an intra-radar instance discrimination learning problem.
Specifically as shown in the radar branch of Fig.~\ref{fig:overall}, for each radar data point $r$, we (1) stocastically obtain two positive views of $r$ using transformations drawn from $\mathcal{T}$, i.e., $t, t^\prime \sim \mathcal{T}$, and (2) encode, project, and $\ell_2$-normalise the positive views as $z_r = g_{\phi^r}(f_{\theta^r}(t(r))), z^{\prime}_r = g_{\phi^r}(f_{\theta^r}(t^{\prime}(r)))$.
Using a mini-batch of $B$ samples, we then compute a contrastive loss~\cite{hadsell2006dimensionality,oord2018representation} for the encoded positive views of the $i$th sample $z_{r, i}$ and $z^{\prime}_{r, i}$ against a set of negative views drawn from the mini-batch:
\begin{align}
  \ell_i^{r \rightarrow r^\prime} = - \log \frac{ \exp({\operatorname{sim}(z_{r, i}, z^{\prime}_{r, i})}) }{ \sum\nolimits^{B}_{j = 0} \exp({\operatorname{sim}(z_{r, i}, z^{\prime}_{r, j}})) } 
  \label{eq:radar-to-radar_loss}
\end{align}
where $\operatorname{sim}(x, y) \coloneqq x^\top y/ \tau$ is a similarity function and $\tau$ is a temperature hyper-parameter.
Similarly, the encoded augmented views can be used as contrastive negatives for added efficiency, which gives us the in-batch symmetric~\cite{chen2020simple} intra-radar loss function.
\begin{align}
   \mathcal{L}_{\text{intra}} &= \frac{1}{2B} \sum\nolimits_i^B( \ell_i^{r \rightarrow r^\prime} + \ell_i^{r^\prime \rightarrow r} )
   \label{eq:intra-radar_loss}
\end{align}


\subsection{Cross-modal radar-vision learning}
\label{sec:cross}
As illustrated in Fig.~\ref{fig:overall}, cross-modal learning uses radar and vision within a joint embedding architecture.
Within this architecture, the pretrained vision model teaches the radar model how to sense and featurise the environment. 
Vision captures visual features from the scene in front of the vehicle. 
Radar data, on the other hand, is preprocessed to create 2D range-azimuth heatmaps, which represent the scene from a BEV perspective. 
While radar and vision operate within these different coordinate systems, their embeddings are nonetheless \emph{aligned} via the contrastive loss.

To implement cross-modal learning, we obtain a prototype radar vector as an average of the two positive vectors $\bar{z}_r = ( z_r + z^\prime_r )/2$ following~\cite{afham2022CrossPoint}. 
We encode and normalize the corresponding vision sample
$z_v = f^\ast_{\theta^v}(v)$. We found it empirically beneficial to omit the MLP projector head from the frozen vision branch while keeping a projector after the radar encoder.

Similar to the radar-to-radar contrastive learning term in Eq.~\ref{eq:radar-to-radar_loss}, we then compute the term $\ell_i^{\bar{r} \rightarrow v}$, where the use of the prototype $\bar{z}_r$ in radar-to-vision contrastive term is denoted by $ \bar{r}$.
The in-batch cross-modal contrastive loss is then given by 
\begin{align}
    \mathcal{L}_{\text{cross}} &= \frac{1}{B} \sum\nolimits_i^B \ell_i^{\bar{r} \rightarrow v}
   \label{eq:cross-modal_loss}
\end{align}

With the intra-modal and cross-modal losses defined in Eqs.~\ref{eq:intra-radar_loss}~\&~\ref{eq:cross-modal_loss}, the overall composite loss is 
\begin{align}
    \mathcal{L} &= \lambda_{\text{intra}} \mathcal{L}_{\text{intra}} +  \mathcal{L}_{\text{cross}}
   \label{eq:composite_loss}
\end{align}
where $\lambda_{\text{intra}}$ is a hyper-parameter.

\subsection{Augmentations}
\label{sec:augs}
\begin{figure}[t]
    \begin{center}
        \includegraphics[width=\columnwidth]{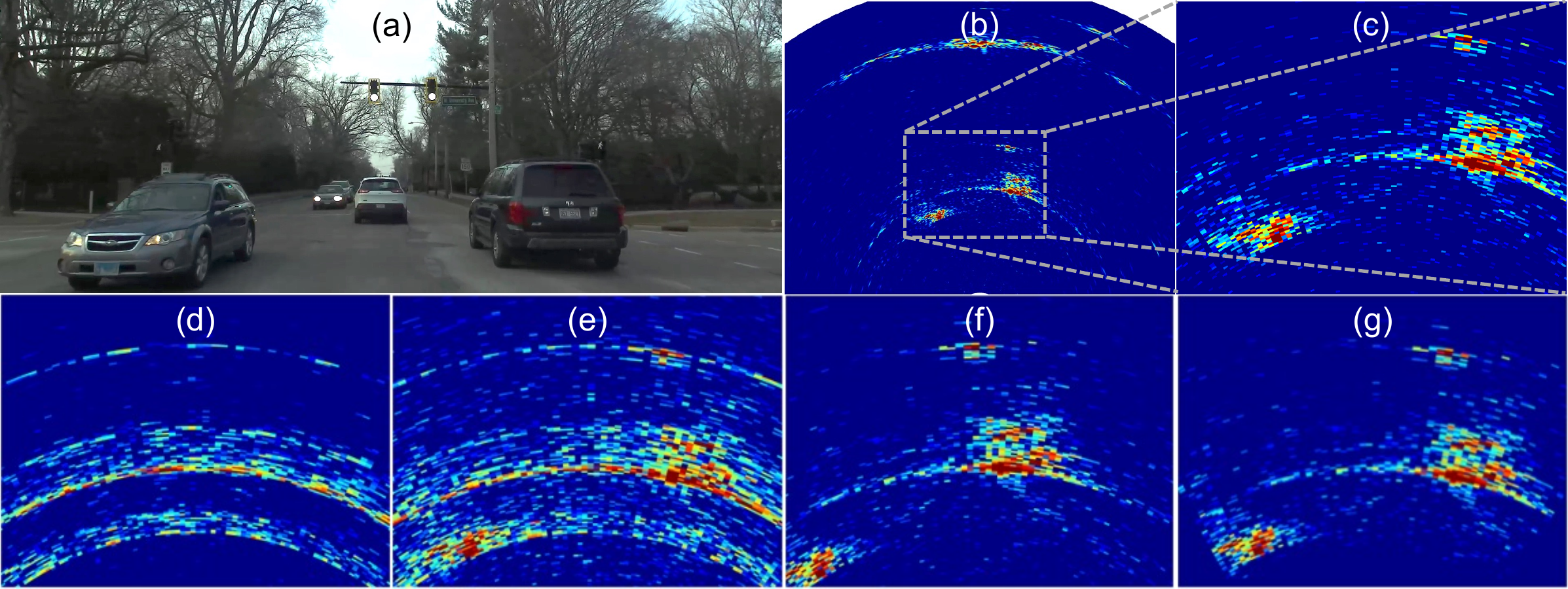}
        \vskip -0.1in
        \caption{\textbf{Radar-specific augmentations.} (a) Scene. (b) Original radar heatmap. (c) Zoomed-in region of cars. (d) Random Phase. (e) Antenna Dropout. (f) Rotation (Polar). (g) Center Cropping (Polar).}
        \label{fig:augmentations}
    \end{center}
    \vskip -0.33in
\end{figure}

A suite of augmentations is essential to our \name\ framework.
We next treat these augmentations, as used in both intra- and cross-modal learning.
We extensively compare and ablate their effectiveness in Sec.~\ref{sec:exp}.
Fig.~\ref{fig:augmentations} gives a visual intuition for all the augmentations we utilize in~\name. 

\subsubsection{Repurposed vision augmentations}
Considering that BEV radar heatmaps have formats similar to camera images, a subset of standard SSL vision augmentations is potentially applicable to radar heatmaps. However, due to the different perspectives and coordinate system, most vision augmentations are not applicable or need to be carefully modified. 

We conduct extensive experiments on different vision augmentations and their combinations (cf., Sec.~\ref{sec:results}). We find that horizontal flip, rotation, and center cropping~\cite{chen2020simple} are also suitable for radar heatmaps. 
We note that for radar heatmaps whose coordinates are polar, rotation and center cropping should be applied in the polar coordinates, as shown in Fig.~\ref{fig:augmentations}(f) and (g) respectively.






\subsubsection{Radar-specific augmentations}
In addition to the repurposed subset of vision augmentations, we introduce and experiment with a new domain-specific augmentation for radar SSL we call Radar MIMO Mask (RMM). We briefly explain how the raw data is processed before RMM is applied.

\vskip 0.05in
\noindent\textbf{RMM implementation.}
Several radar formats typically appear in related work: range-azimuth heatmaps, point clouds, or range-Doppler maps~\cite{qian2021MVDNet, schumann2018semantic, madani2022radatron}.
Differently, \name~uses an intermediate 3-D tensor in order to apply RMM augmentations. 
Specifically, consider a MIMO radar with $M$ transmitters and $N$ receivers. 
A range-azimuth heatmap $r(\rho, \phi) \in \mathbb{R}^{L \times A}$ is generated when a preceding 3-D complex tensor $S \in \mathbb{C}^{MN \times L \times A}$ is integrated noncoherently over all the antenna pairs (at the first index). 
RMM is applied before this integration. 
RMM is best presented as the composition of two operations: (1) antenna dropout,  and (2) random phase noise.
We further explain these below.

\vskip 0.05in
\noindent\textit{(1) Antenna Dropout}.
We leverage the reconfigurability of the virtual array emulated by the MIMO radar to design this radar-specific augmentation.
We omit randomly a subset of virtual antenna elements from subsequent signal aggregation. 
Mathematically, we can write
\begin{align*}
    r^{\prime}(\rho, \phi) &= \left|\sum_{k = 1}^{MN} b_{k} S(\rho, \phi, k) \right|,\quad b_{k} \sim \text{Bernoulli}(p)
\end{align*}
where $r^{\prime}(\rho, \phi)$ is the augmented radar heatmap as a function of range $\rho$ and azimuth angle $\phi$, $k$ indexes the set of $M \times N$ antenna pairs, and $b_{k}$ are independent discrete random masks that nullifies the $k\text{th}$ antenna pair with probability $p$.
This augmentation simulates scenarios with partial sensor failure or obstructions, which promotes learning from incomplete data and improves robustness. 
The probability of antenna dropout $p \in [0, 1]$ is a tunable hyper-parameter. 

\vskip 0.05in
\noindent\textit{(2) Random Phase Noise}.
This augmentation randomizes the phase of the received (complex) signals before their aggregation. 
Mathematically, we can describe this phase randomization as
\begin{align*}
    S^{\prime}_{k} = S_{k} \cdot e^{i\theta_{k}}, \hspace{0.45cm} \theta_{k} \sim U[-\alpha\pi, \alpha\pi), \hspace{0.45cm} 1 \leq k \leq MN
\end{align*}
where $S_k$ is the signal from the $k\text{th}$ transmitter and receiver pair, $S^{\prime}_{k}$ is the augmented signal, and $\theta_{k}$ are i.i.d. phase shifts drawn from uniform distributions $\in [-\alpha\pi, \alpha\pi]$ (in radians), where $\alpha \in [0, 1)$ is a tunable hyper-parameter.
This randomization mimics the phase variability introduced by environmental factors and relative motions between the radar and the scene, which is also referred to as {\it Doppler-induced phase noise}~\cite{madani2022radatron,guan2023exploit}.
Thus it enhances the training coverage of RF conditions likely to occur in the real-world. We note that larger $\alpha$ corresponds to more aggressive movements and noise in the environment.

\vskip 0.05in
\noindent\textbf{RMM instantiation}. 
The final RMM augmentation is the (order-invariant) composition of the two operations detailed above. 
We found empirically that the hyper-parameters $p=0.9$ and $\alpha=0.1$ lead to the best performance in our experiments (see Sec.~\ref{sec:ablation}).


\subsection{Downstream fine-tuning}
After pre-training, we discard the projector head and use the radar backbone only to perform downstream tasks.
We fine-tune the radar backbone with a task-specific head on top. 
Specifically, we demonstrate \name~on the challenging task of bounding box detection for cars using standalone radar heatmaps. 
This task showcases the practical utility of our pre-training towards extending current self-driving perception stacks with weather-immune, fine-grained radar capabilities.

\subsection{Implementation details}
\label{sec:imp}
For the radar backbone, we use \Radatron~\cite{madani2022radatron}, which adopts an FPN-based architecture. 
The backbone has a two-stream architecture, which takes as inputs high- and low-resolution radar heatmaps. 
Specifically, each stream goes through a stem layer and then two ResNet stages, which are identical to the building blocks of ResNet50~\cite{he2016deep}. 
Then the two streams are concatenated and fused in a convolutional layer. 
The resultant feature maps are further encoded via additional ResNet stages, and combined to create the features similar to Detectron2~\cite{wu2019detectron2}. We pre-train the backbone of the model (without the FPN and the linear regression heads) as the radar feature extractor. Future research could benefit from changing the backbone\cite{vit2021, pu2023rotated} and detector\cite{detr2020, hdetr2022, pu2023rank} architectures.

The vision branch uses a pre-trained CLIP image encoder model~\cite{radford2021learning}, which we freeze throughout pre-training.

\section{Experiments and Evaluation}
\label{sec:exp}

\subsection{Dataset}
\label{sec:dataset}
We evaluate \name~on the \Radatron\cite{madani2022radatron} dataset, which supports raw radar data format. 
This is because our domain-specific augmentations require raw radar format.
In addition to the requisite raw radar format, we find Radatron's size beneficial in the characterisation we present herein.
Out of the unlabeled set, we use $32$K frames for self-supervised pretraining, $13$K annotated frames for supervised fine-tuning, and $3$K annotated frames for testing. 
The train and test splits are constructed from experiments conducted on different days throughout the data collection campaign. 
The raw radar frames are first converted to complex, 86-channel heatmaps.
These heatmaps are then fed to the network for preprocessing and stochastic augmentation.

\subsection{Experiments}
\label{sec:exps}
We pre-train \name~as depicted in Fig.~\ref{fig:overall} and detailed in Secs.~\ref{sec:technical}. 
We utilize unlabeled radar-vision frames from \Radatron\ as described in Sec.~\ref{sec:dataset}.
We specialize the pre-trained radar embeddings for a downstream task relevant to self-driving.
The task strives to detect rotated 2D bounding boxes in BEV from radar heatmaps.

During pre-training, we use a batch size of 64, learning rate of 0.05 and cosine learning rate scheduling with an SGD optimizer with momentum 0.9 and weight decay 0.0001. During fine-tuning, we adopt the same training setting as Radatron, using a batch size of 8, an SGD optimizer with learning rate of 0.01 and 25K iterations with learning rate drop at 15K and 20K iterations. We increase the weight decay to 0.001 in order to avoid overfitting problems, boosting the baseline performance.

Unless otherwise stated, our results are obtained with pre-training the backbone using 32K unlabeled frames, and fine-tuning the downstream model on 13K labeled frames. Results are averaged over 6 runs.

\begin{table}[t!]
\begin{minipage}[t]{\linewidth}
    \centering
    \setlength{\tabcolsep}{8pt}
    \resizebox{\linewidth}{!}
    {
        \begin{tabular}{l|ccccc}
           Method & $\operatorname{mAP}$ & $\operatorname{AP}_{75}$ & $\operatorname{AP}_{50}$ \\
            \hline
           \Radatron~\cite{madani2022radatron}& $56.5\pm0.2$  & $64.5\pm1.7$ & $88.9\pm0.4$\\ 
           Intra-modal & $59.4\pm1.0$ & $66.8\pm1.8$ & $89.1\pm0.5$ \\ 
           Cross-modal  & $59.7\pm0.2$  & $67.1\pm0.4$ & $89.3\pm0.3$\\ 
\cellcolor{gray!15!white}\name\ (ours) & \cellcolor{gray!15!white}$62.3\pm0.6$  & \cellcolor{gray!15!white}$69.7\pm1.2$ & \cellcolor{gray!15!white}$89.6\pm0.1$\\ 
        \end{tabular}
    }
    \vskip -0.1in
    \caption{\textbf{Performance of downstream bounding box detection against baselines.} Best performing model is \colorbox{gray!15!white}{highlighted}.}
    \label{tab:exp_radatron_ft}
\end{minipage}
\vskip -0.2in
\end{table}

\subsection{Baselines} \label{sec:baselines}
We evaluate against supervised learning as well as different variants of self-supervised learning in order to expose the merit of our design choices. 
We denote contrastive learning by CL below.
\vskip 0.04in

\noindent\textbf{(1) Radatron.}
We compare against the original implementation reported in~\cite{madani2022radatron} based on supervised learning.

\noindent\textbf{(2) Intra-modal CL.}
We disable vision from contributing to the composite contrastive loss, which results in intra-modal, radar-only CL. For this, we use the vision-based augmentations of vertical flipping and center cropping.

\noindent\textbf{(3) Cross-modal CL.}
We disable intra-modal CL and its radar-specific augmentations, reverting to a CL configuration that is wholly reliant on cross-modal learning between radar and vision. We extend the implementations of SimCLR~\cite{chen2020simple} and MoCo~\cite{he2020momentum} for our cross-modal settings. 
\begin{table}[t!]
\begin{minipage}[t]{\linewidth}
    \centering
    \setlength{\tabcolsep}{8pt}
    \resizebox{\linewidth}{!}
    {
        \begin{tabular}{l|ccccc}
           Method & $\operatorname{mAP}$ & $\operatorname{AP}_{75}$ & $\operatorname{AP}_{50}$ \\
            \hline
           \Radatron~\cite{madani2022radatron} & $22.1\pm0.8$  & $17.7\pm1.2$ & $48.0\pm1.5$ \\ 
           Intra-modal & $45.4 \pm 0.1$ & $48.6\pm0.5$ & $78.3\pm0.1$ \\ 
           Cross-modal & $46.3\pm0.1$  & $49.4\pm0.3$ & $83.0\pm0.1$\\ 
           \cellcolor{gray!15!white}\name\ (ours) & \cellcolor{gray!15!white}$52.6\pm0.1$  & \cellcolor{gray!15!white}$58.5\pm0.2$ & \cellcolor{gray!15!white}$86.9\pm0.1$\\ 
        \end{tabular}
    }
    \vskip -0.1in
    \caption{\textbf{Performance of downstream bounding box detection with frozen backbone in fine-tuning.} Best performing model is \colorbox{gray!15!white}{highlighted}. Results are averaged over 2 runs.}
    \label{tab:exp_radatron_ft_frozen}
\end{minipage}
\vskip -0.2in
\end{table}
\section{Results}
\label{sec:results}
This section presents a comprehensive analysis of \name's performance against baselines and examines the impact of various augmentations on model performance. 

\vskip 0.04in 
\noindent\textbf{Evaluation metrics.} 
Following previous radar detection work~\cite{qian2021MVDNet, madani2022radatron}, we use Average Precision (AP) with IoU thresholds of $0.5$, and $0.75$ to evaluate \name's detection performance.
We also use the mean AP (mAP) of IoU values from 0.5 to 0.95 in 0.05 steps. We follow the COCO framework \cite{lin2014microsoft} for evaluating \name.

\subsection{Performance vs. baselines}

We characterize \name's performance against the baselines enumerated in Sec.~\ref{sec:baselines} and on the downstream task discussed therein. 
Specifically, we analyze performance by means of: (a) fine-tuning the backbone along with the task-specific head, and (b) freezing the backbone and training the task-specific head only.

\vskip 0.05in
\noindent\textbf{Fine-tuning backbone.} 
We pre-train \name~utilising our composite intra- and cross-modal CL configuration, along with the two baseline CL configurations from Sec.~\ref{sec:baselines}. 
We then fine-tune all pre-trained backbones along with their bounding box estimation heads.
We compare these pre-trained variants to the implementation of~\cite{madani2022radatron} which uses random initialization. 
Table~\ref{tab:exp_radatron_ft} shows the quantitative results using three metrics: mAP, AP$_\text{50}$, and AP$_\text{75}$. 
The mean and standard deviation of these results are obtained from 6 different runs of supervised training, while keeping the pre-trained weights the same. 
We see that \name's composite intra- and cross-modal configuration performs most favourably, and outperforms random initialization by 5.8\% in mAP.
This demonstrates the efficacy of \name's pre-training on this highly relevant downstream task.
\name~also outperforms intra-modal CL and cross-modal CL baselines by 2.9\% and 2.4\% respectively.
Despite good gains over random initialisation (approx. 3\% each), the two CL baselines are unable to approach the performance of \name's composite CL loss.

\vskip 0.05in
\noindent\textbf{Freezing backbone.}
We freeze the pre-trained weights in order to assess and compare the quality of the learnt features across our contrastive configurations.
To this end, we train task-specific heads for our downstream bounding box estimation task similar to above.
For Radatron, we randomly initialize its backbone and then similarly freeze it. 
The averages and standard deviations are  listed in Table~\ref{tab:exp_radatron_ft_frozen}. 
We observe that \name~outperforms all baselines on all metrics. 
Random initialisation performs poorly compared to the pre-trained variants.
This highlights the inadequacy of the task-specific head to perform accurate bounding box estimation without quality featurisation underneath. 
We also observe that the gap between \name~and the two CL baselines widens compared to Table~\ref{tab:exp_radatron_ft}. 
Without fine-tuning to compensate, this further underscores the efficacy of~\name~compared to the CL baselines.
We also observe a slight performance advantage to cross-modal over the intra-modal CL. 
This could point to the importance of visual priors in the training of quality radar embeddings.
\vskip 0.05in

\begin{table}[t!]
\begin{minipage}[t]{\linewidth}
    \centering
    \setlength{\tabcolsep}{8pt}
    \resizebox{0.8\linewidth}{!}
    {
        \begin{tabular}{l|ccccc}
           Labeled data fraction & $100\%$ & $10\%$ & $1\%$ \\
            \hline
            \Radatron~\cite{madani2022radatron} & $56.5$ & $42.8$ & $27.9$ \\ 
           \name\ (ours) & $62.3$ & $50.2$ & $39.1$ \\ 
           \hline
           Gain  & {\color{ForestGreen}$+5.8$} & {\color{ForestGreen}$+7.4$} & {\color{ForestGreen}$+11.5$} 
        \end{tabular}
    }
    \vskip -0.1in
    \caption{\textbf{Label efficiency for fine-tuning.} We use all unlabeled data for self-supervised pre-training, and vary size of labeled data for fine-tuning. We use $\operatorname{mAP}$ as our metric.}
    \label{tab:label_density_ft}
\end{minipage}
\vskip -0.18in
\end{table}

\begin{table}[t]
\begin{minipage}[t]{\linewidth}
    \centering
    \setlength{\tabcolsep}{8pt}
    \resizebox{\linewidth}{!}
    {
        \begin{tabular}{l|cccccccc}
           Method & $\operatorname{mAP}$ & $\operatorname{AP}_{75}$ & $\operatorname{AP}_{50}$ \\
            \hline
            RMM & $61.2 \pm 0.5$  & $68.1\pm0.5$ & $89.4\pm0.4$\\
            Rotation  & $61.4 \pm 0.6$  & $68.7\pm1.0$ & $89.0\pm0.6$\\
            Center Cropping & $61.0 \pm 1.0$ & $67.7\pm1.4$ & $89.2\pm0.7$ \\
            Horizontal Flip & $59.6 \pm 0.8$  & $67.5\pm1.3$ & $89.0\pm0.3$\\
            \hline
            Cross-modal (No Aug.) & $60.1 \pm 1.0$  & $67.1 \pm 0.7$ & $89.2 \pm 0.2$\\
            \hline
            Threshold & $59.5 \pm 0.6$  & $66.4\pm1.1$ & $88.7\pm0.5$\\
            Cutout & $58.4 \pm 1.2$ & $66.8\pm1.2$ & $89.0\pm0.5$ \\
            Vertical Flip & $58.1 \pm 0.6$  & $66.6\pm0.9$ & $88.5\pm0.3$\\
        \end{tabular}
    }
    \vskip -0.1in
    \caption{\textbf{Effect of adding one augmentation at a time to the base~\name~net.}}
    \label{tab:ablate:grad_add}
\end{minipage}
\vskip -0.25in
\end{table}


\vskip 0.05in
\noindent\textbf{Label efficiency}. 
We investigate the impact of the number of available labeled data on performance under a fine-tuning protocol.
We compare \name~to the random initialization of Radatron. 
Table~\ref{tab:label_density_ft} shows the mAP after full fine-tuning as a function of the fraction of labeled data used. 
We see increasing improvements using \name~pre-training over the supervised baseline across decreased label density regimes.

\subsection{Ablating augmentations}
\label{sec:ablation}
To better understand the value of each augmentation, we dissect the contribution of individual repurposed and radar-specific augmentations, as well as effect of removing the augmentations from the best combinations. 

\vskip 0.05in 
\noindent\textbf{Individual augmentations}. 
We first compare the effect of adding individual augmentations described in Sec.~\ref{sec:augs}. We also list three augmentations that we experimented with but did not yield beneficial results. They include two standard SSL vision augmentations: Cutout and Vertical Flip~\cite{chen2020simple}. We also tested another radar-specific augmentation, Thresholding, whereby we created a binary mask by setting a power (pixel-value) threshold for the radar heatmap.  

\newcommand{\cmark}{\ding{51}}%
\newcommand{\xmark}{\ding{55}}%

\begin{table}[t]
\begin{minipage}[t]{\linewidth}

\centering
\setlength{\tabcolsep}{8pt}
    \resizebox{\linewidth}{!}
    {
\begin{tabular}{cccc|ccc}
   RMM & CC & HF & ROT & $\operatorname{mAP}$ & $\operatorname{AP}_{75}$ & $\operatorname{AP}_{50}$ \\
    \hline
   \cmark & \cmark & \cmark & \cmark & $61.6 \pm 0.7$  & $68.9 \pm 1.1$ & $89.5 \pm 0.5$ \\
   \xmark & \cmark & \cmark & \cmark & $61.3 \pm 0.8$  & $68.7 \pm 1.3$ & $89.4 \pm 0.2$ \\
   \cmark & \xmark & \cmark & \cmark & $61.0 \pm 0.8$ & $69.1 \pm 1.1$ & $89.5 \pm 0.5$ \\
   \cmark & \cmark & \xmark & \cmark & $61.7 \pm 0.2$  & $68.9 \pm 0.3$ & \cellcolor{gray!15!white}$89.8 \pm 0.5$\\
   \cmark & \cmark & \cmark & \xmark & \cellcolor{gray!15!white}$62.3\pm0.6$  & \cellcolor{gray!15!white}$69.7\pm1.2$ & $89.6\pm0.1$\\
\end{tabular}
}
\vskip -0.1in
\caption{\textbf{Effect of removing each augmentation individually from the four best augmentations found in Table.~\ref{tab:ablate:grad_add}.}}
\label{tab:ablate:grad_remove}
\end{minipage}
\vskip -0.15in
\end{table}

\begin{table}[t]
\begin{minipage}[t]{\linewidth}

\centering
\setlength{\tabcolsep}{8pt}
    \resizebox{0.95\linewidth}{!}
    {
\begin{tabular}{cc|ccc}
    $p$ & $\alpha$ & $\operatorname{mAP}$ & $\operatorname{AP}_{75}$ & $\operatorname{AP}_{50}$ \\
    \hline
    No Aug. & No Aug. & $60.1 \pm 1.0$  & $67.1 \pm 0.7$ & $89.2 \pm 0.2$\\
   $1.0$ & $0.3$ & $61.0 \pm 0.6$  & $67.8 \pm 1.2$ & $89.3 \pm 0.1$\\
   $0.9$ & $0.1$ & \cellcolor{gray!15!white}$61.2 \pm 0.5$  & \cellcolor{gray!15!white}$68.1 \pm 0.5$ & \cellcolor{gray!15!white}$89.4 \pm 0.4$\\
   $0.9$ & $0.3$ & $60.7 \pm 0.5$ & $66.8 \pm 0.9$ & $89.0 \pm 0.5$ \\
   $0.9$ & $0.5$ & $60.0 \pm 0.7$  & $67.3 \pm 0.8$ & $89.1 \pm 0.2$\\
   $0.8$ & $0.3$ & $60.3 \pm 0.5$  & $67.3 \pm 0.9$ & $88.9 \pm 0.8$\\
\end{tabular}
}
\vskip -0.1in
\caption{\textbf{Hyper-parameters of the RMM augmentation.}}
\label{tab:ablate_rw}
\end{minipage}
\vskip -0.25in
\end{table}

\begin{figure*}[t!]
	\begin{center}
		\includegraphics[width=\linewidth]{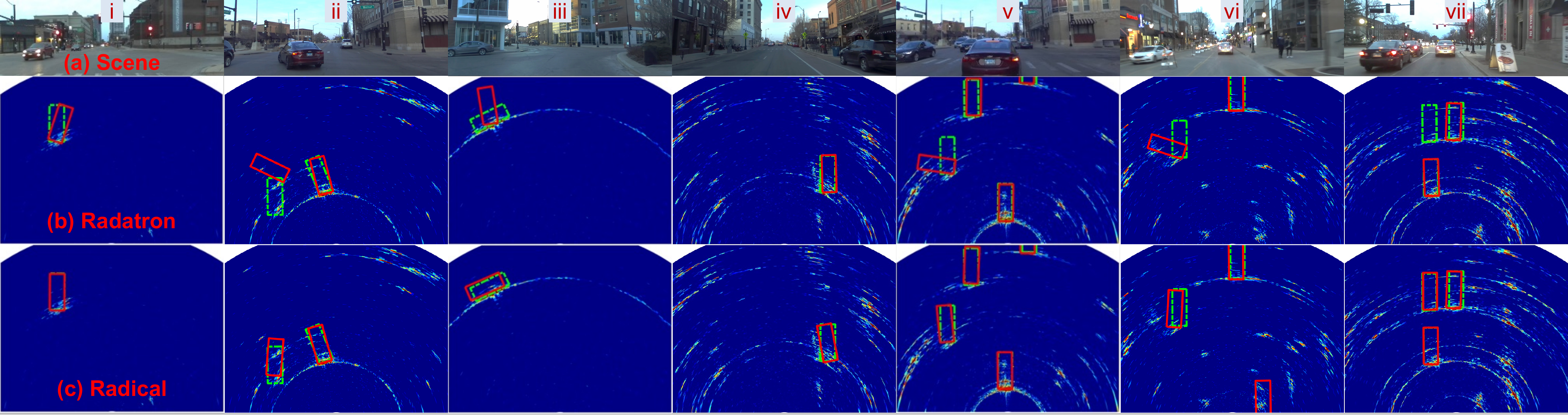}
		\vskip -0.1in
		\caption{\textbf{Examples from our test set:} (a) Original scene. (b) \Radatron\ (supervised) baseline. (c) \name. Groundtruth marked in green and predictions in red.}
		\label{fig:qualitative}
	\end{center}
	\vskip -0.35in
\end{figure*}

In our experiments, we pre-train the \name~net using one augmentation at a time, and fine-tune it with the 13K labeled dataset. The results are shown in Table~\ref{tab:ablate:grad_add}. As a baseline for comparison, we also include the cross-modal only baseline in the table which does not use any augmentations. As seen, four out of the seven tested augmentations, namely rotation, RMM, center crop, and horizontal flip prove beneficial for pre-training.
On the other hand, thresholding, cutout, and vertical flipping will on average be detrimental to performance across all three listed metrics. 
We make the following points. 
First, based on these results, we removed the three worst-performing augmentations from our final model. 
Second, we make the following observations regarding the effectiveness of each augmentation:
\begin{enumerate}
\item While radar heatmaps are symmetric along the mid-point of the azimuth axis, they are certainly not so along the range axis. This is why horizontal flip retains the underlying structure of the radar data, while vertical flip fails to do so and hurts performance. 
\item While thresholding might seem an intuitive extension to similar quantization methods in vision, it fails to aid performance in radar data because radar data is already extremely sparse in nature. 
\item Center cropping and rotation borrowed from vision seem to boost \name~performance. They preserve the underlying semantics of radar heatmaps. 
\item RMM is a useful MIMO radar-specific augmentation.
\end{enumerate}

\vskip 0.1in
\noindent\textbf{Combining augmentations.} 
Having found four individually useful augmentation in Table.~\ref{tab:ablate:grad_add}, we next explore how to best combine them.
To this end, we conduct five experiments; the first uses all four augmentations, while each subsequent experiments removes one out of four augmentations at a time. 
The results are shown in Table.~\ref{tab:ablate:grad_remove}. 
We note that all these combinations perform seemingly equally well under the AP$_{50}$ metric. 
However, metrics mAP and AP$_{75}$ reveal that the combination RMM + Center Crop + Horizontal Flip is the clear winner.
We hence use it in \name's final model.
\vskip 0.05in
\noindent\textbf{Hyper-parameters of RMM augmentation}. 
Having just introduced RMM augmentation in this work, we next explore the  configuration space of its hyper-parameters in order to identify an initial performant recipe.
Table~\ref{tab:ablate_rw} sweeps $p$ and $\alpha$ (cf. Sec.~\ref{sec:augs}). 
To establish a comparative baseline, the first row of Table~\ref{tab:ablate_rw} shows our three AP metrics without using RMM.
We observe that keeping virtual antennas with probability $p=0.9$ and noise randomisation with $\alpha=0.1$ results in best performance across the three metrics. 
This amounts to randomly omitting 10\% of the antennas. 
While sizeable, we view this antenna masking as non-aggressive and preserving of the integrity of the radar data.

\subsection{Qualitative results}
\label{sec:qualitative}
We next present qualitative results and compare~\name~to the supervised baseline Radatron.
Fig.~\ref{fig:qualitative} shows groundtruth as dotted green bounding boxes, and model predictions in solid red.
Fig.~\ref{fig:qualitative} consists of three rows: (1) upper row depicts front-view camera images, (2) middle row depicts Radatron's bounding box predictions overlaid on top of groundtruth in BEV, and (3) bottom row depicts \name's bounding box predictions and groundtruth.
We make the following observations.
First, quite a few of the failure cases---namely columns i, ii, iii, v, and vi in Fig.~\ref{fig:qualitative}---in the baseline arise from scenarios where a car is detected, albeit its orientation and exact bounding box are missed. 
This is due to the low-resolution and specular nature of radar. 
These failures are mostly rectified by \name's network as seen in the bottom row. Second, \name\ performs better in scenarios where a car's radar reflection might get occluded by other cars in the scene, as in Fig.~\ref{fig:qualitative}(vii). Both these failure cases are well known in radar object detection systems, as shown by previous work~\cite{guan2020through, madani2022radatron}. \name\ overcomes these failures thanks to pre-training radars to learn radar priors like specularity and sparsity jointly with vision features, which additionally carry semantic information such as precise car location and orientations. Finally, we note that Radatron performs reasonably closer compared to \name\ when detecting the approximate location of vehicles which is reflective of their relatively closer AP$_{50}$ performance compared to mAP and AP$_{75}$. 
In other words, \name's strength, as noted in Sec.~\ref{sec:baselines}, lies in its more precise box detection in complex situations, illustrated in Fig.\ref{fig:qualitative}.

\begin{figure}[t!]
\begin{center}
    \includegraphics[width=\linewidth]{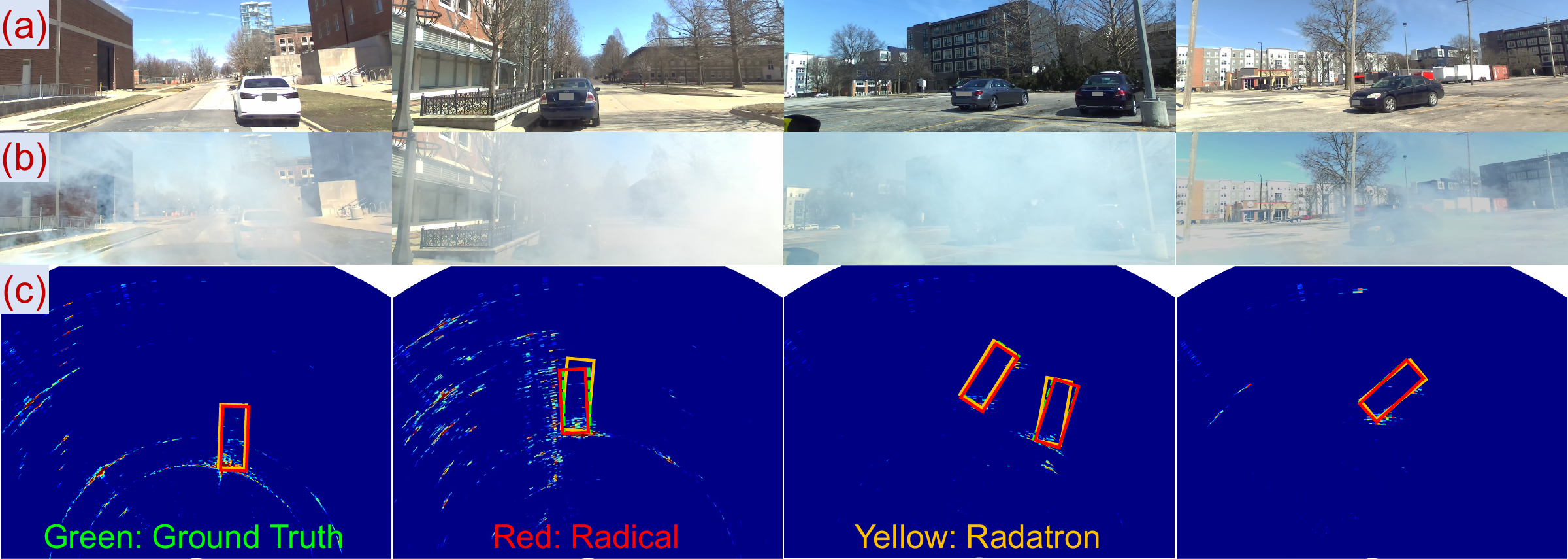}
    \vskip -0.1in
    \caption{\textbf{Controlled Fog Experiment.} (a) Scene. (b) Scene in fog. (c) Prediction overlaid on radar heatmap captured in fog.}
    \label{fig:fog}
\end{center}
\vskip -0.4in
\end{figure}

\noindent\textbf{Controlled Fog Experiment}. We evaluate \name\ on fog data in \Radatron\ dataset. Figure~\ref{fig:fog} shows \name\ can detect cars accurately in fog and even outperform \Radatron.

\section{Conclusion}
\label{sec:conclusion}
In this paper, we presented a self-supervised approach to radar object detection in the context of self-driving cars, harnessing the largely untapped potential of vast quantities of unlabeled radar data. Our extensive evaluations illustrate that \name\ achieves superior performance over supervised baselines by effectively combining intra- and cross-modal self-supervised learning, and employing radar-specific as well as vision-inspired augmentations in the context of contrastive learning. It is our hope that these contributions are followed by future advancements in the field of automotive radar.
{
    \small
    \bibliographystyle{ieeenat_fullname}
    \bibliography{./reference}
}


\end{document}